\PassOptionsToPackage{table}{xcolor}
\documentclass[letterpaper]{article} % DO NOT CHANGE THIS
\usepackage{aaai2027}  % DO NOT CHANGE THIS
\usepackage[hyphens]{url}  % DO NOT CHANGE THIS
\usepackage{graphicx} % DO NOT CHANGE THIS
\usepackage{natbib}  % DO NOT CHANGE THIS AND DO NOT ADD ANY OPTIONS TO IT
\usepackage{caption} % DO NOT CHANGE THIS AND DO NOT ADD ANY OPTIONS TO IT
\usepackage{algorithm}
\usepackage{algorithmic}

\usepackage{booktabs}

\usepackage{xcolor}
\usepackage{amsmath}
\usepackage{amssymb}
\usepackage{multirow}
\usepackage{makecell}
\usepackage{array}
\usepackage{tabularx}
\usepackage{xspace}

\definecolor{mygreen}{HTML}{008744}
\definecolor{myorange}{HTML}{ffa700}
\definecolor{myred}{HTML}{d62d20}
\definecolor{myblue}{HTML}{0668E1}
\definecolor{lightblue}{HTML}{EAF3FF}
\definecolor{lightgreen}{HTML}{E9F8EF}
\definecolor{lightorange}{HTML}{FFF4E2}
\definecolor{lightred}{HTML}{FFECEC}
\DeclareMathOperator*{\argmax}{arg\,max}
\newcommand{\mc}[1]{\mathcal{#1}}
\newcommand{\tsc}[1]{\textsc{#1}}
\newcommand{\ttt}[1]{\texttt{#1}}
\newcommand{\method}{AuroSFT\xspace}
\newcommand{\msft}{\tsc{mSFT}\xspace}
\newcommand{\lora}{\tsc{LoRA}\xspace}
\newcommand{\aurora}{\tsc{AuroRA}\xspace}

\title{Beyond Full-Model Rollback: AuroSFT for\\Adapter-State Multi-Task Fine-Tuning}
\author{
Yue Han\textsuperscript{\rm 1},
Ziniu Liu\textsuperscript{\rm 2},
Changjian Li\textsuperscript{\rm 2},
Jie Zhang\textsuperscript{\rm 1},\\
Ziyi Chen\textsuperscript{\rm 1},
Tao Wang\textsuperscript{\rm 1},
Yexin Cui\textsuperscript{\rm 2},
Weihong Han\textsuperscript{\rm 3}
}
\affiliations{
\textsuperscript{\rm 1}College of Systems Engineering, National University of Defense Technology\\
\textsuperscript{\rm 2}College of Computer Science and Technology, National University of Defense Technology\\
\textsuperscript{\rm 3}Vernal Institute, China Electronics Corporation\\
\{hanyue, liuzn\_nudt, lichangjian23, zhangjie, wangtao1976, yexincuicyx\}@nudt.edu.cn\\
chenziyi\_nudt@outlook.com,
hanwh@pcl.ac.cn
}

\begin{document}

\maketitle

\begin{abstract}
Multi-task supervised fine-tuning (SFT) often casts a heterogeneous data mixture as a single optimization problem, even though different tasks may reach their best generalization at different times. \msft exposes this mismatch through task-wise roll-out, exclusion, and rollback, but its original formulation materializes the scheduler state as full-model checkpoints, making stage transitions costly to store, restore, and deploy. This paper introduces \method, a parameter-efficient framework that recasts the carried state of overfitting-aware multi-task SFT as a compact, mergeable adapter state. \method freezes the pretrained backbone, trains only injected adapters, rolls back adapter checkpoints at task-wise peaks, and continues on the remaining active mixture. At the layer level, each adapter applies an AuroRA-inspired adaptive nonlinear layer to a low-rank weight factor rather than to the sample representation. The resulting update remains linear in the input, rank-bounded, and exactly mergeable into the frozen projection. Under the retained-backbone comparison protocol, \method achieves 61.36\% average accuracy, compared with 59.85\% for the corresponding \msft reference row, and obtains higher accuracy on all five backbones. Our code is available at the anonymous repository: \url{https://anonymous.4open.science/r/AuroSFT-80D1}.
\end{abstract}

\section{Introduction}

\begin{figure}[t]
    \centering
    \includegraphics[width=\linewidth]{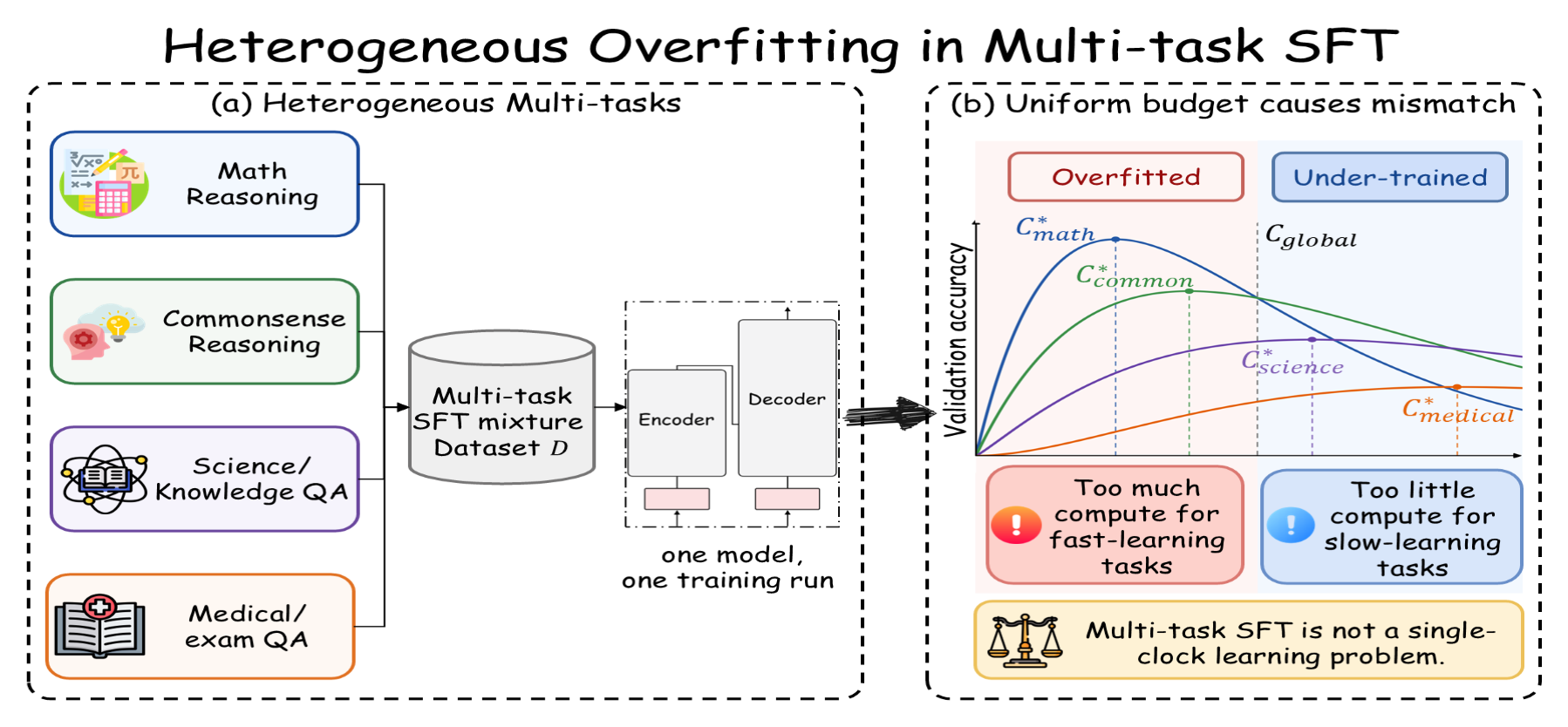}
    \caption{\textbf{Heterogeneous overfitting in multi-task SFT.} A shared global budget can over-train fast tasks while leaving slower tasks under-trained, exposing the need for peak-aware multi-task scheduling.}
    \label{fig:problem}
\end{figure}

Supervised fine-tuning is a standard interface for adapting LLMs to instruction-following skills \citep{zhang2023instructiontuning}, but multi-task SFT rests on a convenient fiction. All tasks can be trained under a single global compute budget. Figure~\ref{fig:problem} shows where this breaks. Math, commonsense, medical QA, and language-understanding tasks may improve under the same optimizer, yet need not peak on the same clock. More training can refine one task while eroding another.

This tension makes homogeneous SFT an awkward interface for heterogeneous supervision. \msft repairs the timing mismatch by rolling out the active mixture, detecting task-wise validation peaks, removing the earliest-overfitted task, rolling back to its peak checkpoint, and continuing on the remaining tasks \citep{koh2026msft}. Its key insight is temporal. Tasks should leave training at different times rather than inherit one global stopping point.

Yet this repair leaves a second design choice hidden in plain sight. Once a task is excluded, \emph{what exactly should be rolled back and carried forward?} In full-model \msft, the answer is the entire model. That answer is simple, but heavy. Every stage transition moves between large checkpoints, tying a data-side schedule to full-parameter state transfer.

Parameter-efficient fine-tuning suggests a different object. Adapters \citep{houlsby2019parameter}, prefix tuning \citep{li2021prefix}, and \lora \citep{hu2021lora} show that adaptation can often be represented by a small trainable state while the pretrained backbone stays fixed. However, these methods are usually placed under a fixed data schedule. They compress the update, but they do not decide how heterogeneous tasks should enter or leave training. Nonlinear low-rank adaptation, such as \aurora, further enriches this adapter state space \citep{dong2025aurora}. For the scheduler studied here, the carried state should be compact, checkpointable, rollback-compatible, and mergeable. This motivates applying the nonlinear mapping to the weight factor rather than to the sample representation.

\method makes this object explicit. We keep the \msft scheduling mechanism, but move its stage state from the full model into a nonlinear weight-transformed low-rank adapter. In each Transformer projection layer, \method freezes the pretrained weight $W_0$ and trains low-rank factors $A$ and $B$. The nonlinear layer transforms the weight factor into $\bar A$ before the input is applied, rather than transforming the sample representation itself. This change is small in notation but central in state space. The scheduler still follows task peaks, but the object it moves through time is compact, adapter-only, and mergeable. The adapted layer remains linear in the input, the update rank is bounded by $r$, and the trained update can be folded back into the frozen projection after training. During the \msft-style schedule, only $\{A,B,\mathrm{ANL}\}$ is optimized, checkpointed, rolled back, and carried across stages. The formal adapter equations are given in the Method section.

From this perspective, the central question is clear. \emph{What state should an overfitting-aware scheduler carry when storage, rollback, and deployment matter?} This paper answers with a mergeable nonlinear low-rank adapter state, rather than a full backbone checkpoint or an input-level nonlinear branch. Matched Qwen2.5-3B ablations guard against a simple ``mSFT+LoRA'' reading. Removing Weight ANL lowers accuracy from 73.30\% to 72.40\%, and removing rollback yields 72.70\%. The main contributions are summarized below.
\begin{itemize}
    \item \textbf{State-space formulation.} Heterogeneous multi-task SFT is framed as \emph{adapter-state transfer under task-wise overfitting}. Peak detection and rollback specify when the training state should move. The proposed formulation specifies what object should move.
    \item \textbf{Weight-transformed adapter.} A nonlinear low-rank adapter is introduced for this state space. Its nonlinear mapping acts on a low-rank weight factor rather than on the input representation, preserving input linearity, rank boundedness, and exact mergeability into the frozen backbone used at inference.
    \item \textbf{AuroSFT framework.} The resulting training procedure freezes the backbone, optimizes only adapter parameters, rolls back compact adapter checkpoints across stages, and preserves the task-exclusion logic of \msft without storing a sequence of full backbones.
    \item \textbf{Empirical validation.} Across ten benchmarks and five lightweight backbones, \method raises the retained-backbone average from 59.85\% for the reported/reproduced \msft reference row to 61.36\% and is higher on all five backbones. Matched Qwen2.5-3B ablations isolate iterative scheduling, rollback, and Weight ANL.
\end{itemize}

\begin{figure*}[t]
\centering
\includegraphics[width=\textwidth]{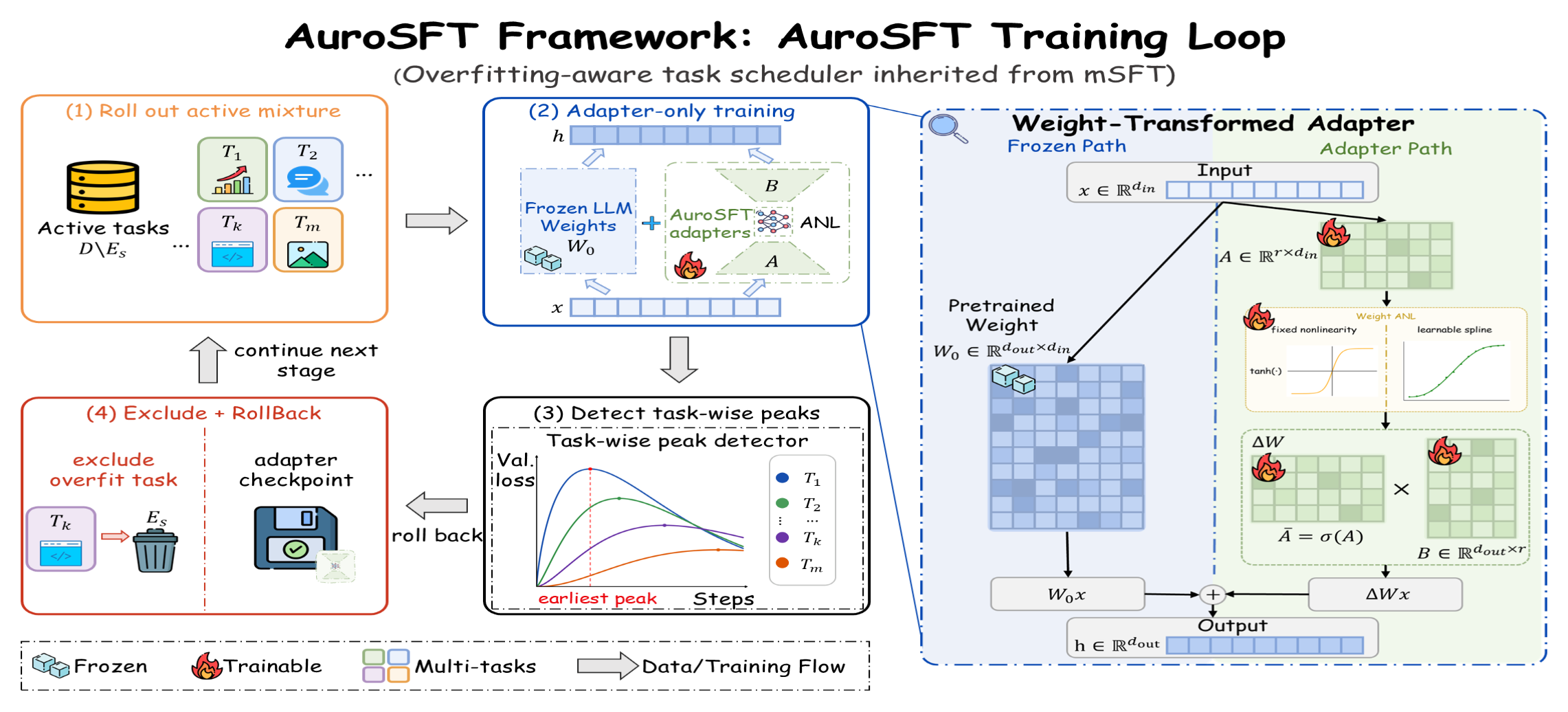}
\caption{\textbf{AuroSFT framework and weight-transformed adapter.} The unified loop couples overfitting-aware task scheduling with adapter-only rollback and mergeable weight-transformed updates.}
\label{fig:overview}
\end{figure*}

\section{Related Work}

\paragraph{Task-State Scheduling for Heterogeneous SFT.}
Modern post-training mixtures combine math, knowledge, dialogue, safety, and reasoning data, making data allocation a first-order design problem for open-weight LLMs \citep{qwen2025qwen25technicalreport,walsh2025,grattafiori2024llama}. Existing work therefore studies example selection, data quantity, mixture weighting, dynamic sampling, data-mixing optimization, and mixup-style supervision \citep{dong-etal-2024-abilities,li-etal-2024-quantity,zhu-etal-2025-dynamic,shi2025damodatamixingoptimizer,wang2026hbo,li2025data,xiao2024sftmix}. Related stopping methods prevent wasted optimization globally, per instance, or through online SFT schedules \citep{PRECHELT1998761,JMLR:v23:21-0983,yuan2025instancedependent,shin2025dynamixsft,zhu-etal-2025-dynamic}. The closest task-level method is \msft, which detects task-wise peaks, excludes the earliest-overfitted task, rolls back, and continues on the active mixture \citep{koh2026msft}. These works identify the right failure mode--heterogeneous tasks do not share one optimal budget--but they mostly optimize the data path or rely on full-model rollback. \method keeps the \msft scheduling signal while changing the carried state. Task peaks trigger transitions in a compact adapter state rather than in the whole backbone.

\paragraph{Adapter State Spaces for Parameter-Efficient Rollback.}
Parameter-efficient tuning replaces full-model updates with smaller trainable objects, including adapters, prefix or prompt parameters, low-rank LoRA factors, and variants that improve memory use, rank allocation, decomposition, or initialization \citep{houlsby2019parameter,li2021prefix,lester2021power,hu2021lora,dettmers2023qlora,zhang2023adalora,liu2024dora,wang2024loraga}. Nonlinear low-rank adaptation further increases adapter expressiveness. \aurora is the closest inspiration, introducing an Adaptive Nonlinear Layer for nonlinear LoRA-style updates \citep{dong2025aurora}. Yet this line usually treats the adapter as a cheaper substitute for full fine-tuning under a fixed training schedule, not as the mutable state of a temporal multi-task algorithm. \method assigns the adapter this second role. Its ANL transforms the low-rank weight factor before the input is applied, so the resulting update remains input-linear, rank-bounded, and mergeable after training. Thus \method is not merely a stronger sampler or a stronger adapter. It couples overfitting-aware task scheduling with a weight-transformed parameter state that can be optimized, checkpointed, rolled back, continued, and deployed without storing a sequence of full backbones.

\section{Method}
\label{sec:method}

Figure~\ref{fig:overview} summarizes the full AuroSFT loop. The left side shows active-task roll-out, peak detection, exclusion, and adapter rollback, while the zoom-in isolates the weight-transformed adapter inside the frozen LLM.

\subsection{Problem Setup}

Let $\mc{D}=\{\mc{D}_1,\ldots,\mc{D}_N\}$ denote a multi-task SFT mixture. A model with parameters $\theta$ is evaluated on held-out sets for each task. Under homogeneous SFT, all tasks share one compute value $c$, even though each task may have its own optimal stopping point:
\begin{equation}
    c_i^* = \argmax_c \; \mathrm{Metric}(\theta_c; \mc{D}_i^{\mathrm{eval}}).
\end{equation}
When $c>c_i^*$, task $i$ may overfit. When $c<c_i^*$, it may remain under-trained. \method preserves this heterogeneous-overfitting view while restricting the trainable state to compact adapters.

\subsection{Overfitting-Aware Scheduler}

\method inherits the iterative roll-out and roll-back scheduler of \msft. At stage $s$, the algorithm trains on the active mixture $\mc{D}\setminus \mc{E}_s$, where $\mc{E}_s$ is the set of excluded tasks. It records task-wise validation metrics, identifies the earliest task peak, excludes that task, and rolls back to the corresponding checkpoint before the next stage begins.
Algorithm~\ref{alg:aurosft} summarizes this stage-wise roll-out, task-exclusion, and roll-back procedure and makes explicit that only the adapter state is restored at a transition. The scheduler itself is not our claimed redesign of \msft. The redesign is the state space in which the scheduler operates. The rolled-back object is the adapter state, not a full model copy of the backbone.

\begin{algorithm}[t]
\caption{\method Scheduler}
\label{alg:aurosft}
\small
\begin{algorithmic}[1]
\STATE \textbf{Input:} Dataset mixture $\mc{D}$, frozen backbone $W_0$, adapter state $\phi_0=\{A,B,\mathrm{ANL}\}$, stage budget $C$
\STATE \textbf{Output:} Final adapter state $\hat{\phi}$ and optionally merged model $W_0+\Delta W$
\STATE $\mc{E}_0\leftarrow\emptyset$; $\hat{\phi}\leftarrow\phi_0$
\FOR{$s=0,1,\ldots$}
    \IF{$\mc{D}\setminus \mc{E}_s=\emptyset$}
        \STATE \textbf{break}
    \ENDIF
    \STATE $\phi(c), \{\mathrm{acc}_i(c)\}_{i,c} \leftarrow \mathrm{AdapterRollout}(W_0,\hat{\phi},\mc{D}\setminus \mc{E}_s,C)$
    \STATE $c_i^* \leftarrow \argmax_c \mathrm{acc}_i(c)$ for each $\mc{D}_i\notin\mc{E}_s$
    \STATE $(c_{\min}, i_{\min}) \leftarrow \arg\min_i c_i^*$
    \IF{$c_{\min}=C$}
        \STATE $\mc{E}_{s+1}\leftarrow \mc{E}_s$
        \STATE $\hat{\phi}\leftarrow \phi(C)$
    \ELSE
        \STATE $\mc{E}_{s+1}\leftarrow \mc{E}_s\cup\{\mc{D}_{i_{\min}}\}$
        \STATE $\hat{\phi}\leftarrow \phi(c_{\min})$ \COMMENT{rollback only adapter state}
    \ENDIF
\ENDFOR
\end{algorithmic}
\end{algorithm}

\subsection{Weight-Transformed Low-Rank Adapter}

For a frozen linear projection $W_0\in\mathbb{R}^{d_{\mathrm{out}}\times d_{\mathrm{in}}}$, \method injects trainable low-rank factors $A\in\mathbb{R}^{r\times d_{\mathrm{in}}}$ and $B\in\mathbb{R}^{d_{\mathrm{out}}\times r}$. Instead of forming the standard LoRA update $BA$ directly, we first transform the weight factor $A$:
\begin{equation}
    \bar A=\sigma(A^\top)^\top.
    \label{eq:weight-transform}
\end{equation}
The adapter update and layer output are:
\begin{equation}
    \Delta W = \frac{\alpha}{r}B\bar A,\qquad
    h = W_0x+\Delta W\,x = (W_0+\Delta W)x.
    \label{eq:delta}
\end{equation}
The ANL $\sigma$ combines a fixed tanh pathway and a learnable spline pathway. Under the row-wise convention used by the implementation, it is written as
\begin{equation}
    \sigma(Z)=\tanh\!\big(\tanh(Z)H^\top\big)+S(Z)W_s^\top,
    \label{eq:anl}
\end{equation}
where $H$ is a trainable self-projection, $S(\cdot)$ denotes B-spline bases, and $W_s$ are learnable spline weights.

This formulation deliberately differs from the input-level \aurora expression in which ANL acts on the projected sample representation. In \method, the input $x$ is multiplied by $\bar A$ only after the weight transformation has been formed. The adapted layer therefore remains linear in $x$, its update rank is at most $r$, and the trained adapter can be merged exactly into the frozen projection by replacing $W_0$ with $W_0+\Delta W$ at deployment.

\subsection{Implementation Alignment}

The implementation freezes the backbone, injects adapters into q/k/v/o/gate/up/down projections, applies ANL to the low-rank weight factor rather than task inputs, and restores only adapter checkpoints at stage transitions.

\setcounter{table}{2}
\begin{table*}[!t]
\centering
\scriptsize
\setlength{\tabcolsep}{2.2pt}
\renewcommand{\arraystretch}{0.95}
\resizebox{0.94\textwidth}{!}{
\begin{tabular}{>{\centering\arraybackslash}m{2.65cm} l *{12}{r}}
\toprule
\multirow{3}{*}{\makecell{\textbf{Benchmark}\\\textbf{Type}}} & \textbf{Model:} & \multicolumn{2}{c}{\textbf{\ttt{OLMo2}}} & \multicolumn{8}{c}{\textbf{\ttt{Qwen2.5}}} & \multicolumn{2}{c}{\multirow{2}{*}{\raisebox{-0.75ex}{\textbf{Average}}}} \\
\cmidrule(lr){3-4} \cmidrule(lr){5-12}
& \textbf{Size:} & \multicolumn{2}{c}{\textbf{\ttt{1B}}} & \multicolumn{2}{c}{\textbf{\ttt{0.5B}}} & \multicolumn{2}{c}{\textbf{\ttt{1.5B}}} & \multicolumn{2}{c}{\textbf{\ttt{3B}}} & \multicolumn{2}{c}{\textbf{\ttt{7B}}} & \multicolumn{2}{c}{} \\
\cmidrule(lr){3-4} \cmidrule(lr){5-6} \cmidrule(lr){7-8} \cmidrule(lr){9-10} \cmidrule(lr){11-12} \cmidrule(lr){13-14}
& \textbf{Metric:} & \textbf{Acc. (\%)} & \textbf{Ep.} & \textbf{Acc. (\%)} & \textbf{Ep.} & \textbf{Acc. (\%)} & \textbf{Ep.} & \textbf{Acc. (\%)} & \textbf{Ep.} & \textbf{Acc. (\%)} & \textbf{Ep.} & \textbf{Acc. (\%)} & \textbf{Ep.} \\
\midrule
\multirow{7}{*}{\makecell{\textbf{Science}\\\textbf{and}\\\textbf{Knowledge}}} & Base & 32.40 & --- & 26.10 & --- & 54.60 & --- & 12.10 & --- & 4.00 & --- & 25.84 & --- \\
& SFT & 47.90 & 9.75 & 37.50 & 0.50 & 65.80 & 3.00 & 70.70 & 5.00 & 74.50 & 2.00 & 59.28 & 4.05 \\
& Continual SFT & 48.50 & 1.90 & 24.60 & 1.95 & 66.60 & 2.08 & 69.80 & 1.80 & 72.90 & 1.40 & 56.48 & 1.83 \\
& DynamixSFT & 47.90 & 5.75 & 39.50 & 0.50 & 65.60 & 2.75 & 70.60 & 3.00 & 74.50 & 7.25 & 59.62 & 3.85 \\
& IES & 47.60 & 10.00 & 39.50 & 0.50 & 65.40 & 4.00 & 70.90 & 3.50 & 74.40 & 3.00 & 59.56 & 4.20 \\
& \msft & 50.40 & 9.75 & 39.20 & 0.25 & 65.40 & 4.75 & 71.50 & 5.50 & 73.60 & 1.50 & 60.02 & 4.35 \\
& \method & 50.88 & 1.75 & 41.62 & 0.25 & 65.50 & 2.00 & 71.75 & 0.50 & 75.00 & 2.50 & 60.95 & 1.40 \\
\midrule
\multirow{7}{*}{\makecell{\textbf{Commonsense}\\\textbf{and}\\\textbf{Language}}} & Base & 9.90 & --- & 22.20 & --- & 42.50 & --- & 8.10 & --- & 8.40 & --- & 18.22 & --- \\
& SFT & 50.90 & 9.75 & 32.90 & 0.50 & 73.00 & 3.00 & 80.40 & 5.00 & 84.20 & 2.00 & 64.28 & 4.05 \\
& Continual SFT & 48.60 & 1.90 & 19.00 & 1.95 & 71.10 & 2.08 & 78.20 & 1.80 & 86.10 & 1.40 & 60.60 & 1.83 \\
& DynamixSFT & 49.00 & 5.75 & 39.90 & 0.50 & 72.60 & 2.75 & 81.90 & 3.00 & 84.60 & 7.25 & 65.60 & 3.85 \\
& IES & 51.00 & 10.00 & 38.80 & 0.50 & 72.60 & 4.00 & 81.70 & 3.50 & 85.50 & 3.00 & 65.92 & 4.20 \\
& \msft & 53.80 & 9.75 & 42.50 & 0.25 & 72.80 & 4.75 & 80.20 & 5.50 & 86.50 & 1.50 & 67.16 & 4.35 \\
& \method & 60.75 & 1.75 & 53.00 & 0.25 & 75.37 & 2.00 & 81.38 & 0.50 & 85.80 & 2.50 & 71.26 & 1.40 \\
\midrule
\multirow{7}{*}{\makecell{\textbf{Mathematic}\\\textbf{and}\\\textbf{Quantitative}}} & Base & 19.50 & --- & 26.20 & --- & 42.80 & --- & 58.00 & --- & 68.00 & --- & 42.90 & --- \\
& SFT & 20.20 & 9.75 & 24.20 & 0.50 & 43.00 & 3.00 & 58.30 & 5.00 & 66.50 & 2.00 & 42.44 & 4.05 \\
& Continual SFT & 18.50 & 1.90 & 23.80 & 1.95 & 45.00 & 2.08 & 58.00 & 1.80 & 67.00 & 1.40 & 42.46 & 1.83 \\
& DynamixSFT & 20.80 & 5.75 & 25.00 & 0.50 & 43.20 & 2.75 & 57.50 & 3.00 & 65.80 & 7.25 & 42.46 & 3.85 \\
& IES & 21.50 & 10.00 & 25.50 & 0.50 & 43.00 & 4.00 & 57.80 & 3.50 & 65.20 & 3.00 & 42.60 & 4.20 \\
& \msft & 23.20 & 9.75 & 23.50 & 0.25 & 48.80 & 4.75 & 62.10 & 5.50 & 70.00 & 1.50 & 45.52 & 4.35 \\
& \method & 18.75 & 1.75 & 19.50 & 0.25 & 47.50 & 2.00 & 60.25 & 0.50 & 66.00 & 2.50 & 42.40 & 1.40 \\
\midrule
\multirow{7}{*}{\makecell{\textbf{Average}\\\textbf{Accuracy}\\\textbf{Across 10}\\\textbf{Benchmarks}}} & Base & 20.80 & --- & 24.60 & --- & 47.40 & --- & 19.70 & --- & 18.60 & --- & 26.22 & --- \\
& SFT & 43.60 & 9.75 & 33.00 & 0.50 & 64.10 & 3.00 & 72.10 & 5.00 & 76.80 & 2.00 & 57.92 & 4.05 \\
& Continual SFT & 42.60 & 1.90 & 22.20 & 1.95 & 64.10 & 2.08 & 70.80 & 1.80 & 77.00 & 1.40 & 55.34 & 1.83 \\
& DynamixSFT & 42.90 & 5.75 & 36.80 & 0.50 & 64.00 & 2.75 & 72.50 & 3.00 & 76.80 & 7.25 & 58.60 & 3.85 \\
& IES & 43.80 & 10.00 & 36.40 & 0.50 & 63.80 & 4.00 & 72.60 & 3.50 & 77.00 & 3.00 & 58.72 & 4.20 \\
& \msft & 46.30 & 9.75 & 37.40 & 0.25 & 65.00 & 4.75 & 73.10 & 5.50 & 77.45 & 2.50 & 59.85 & 4.55 \\
& \method & \textbf{48.40} & 1.75 & \textbf{41.75} & 0.25 & \textbf{65.85} & 2.00 & \textbf{73.30} & 0.50 & \textbf{77.50} & 2.50 & \textbf{61.36} & 1.40 \\
\bottomrule
\end{tabular}
}
\caption{\textbf{Main experimental results.} The table follows the original mSFT task-grouped layout for the five evaluated backbones and adds \method as the proposed adapter-state method.}
\label{tab:main-results}
\end{table*}

\section{Experiments}
\label{sec:experiments}

\subsection{Experimental Setup}

\textbf{Environment and software.}
All audited training runs use an Ubuntu 22.04.4 remote server with two NVIDIA RTX PRO 6000 Blackwell Server Edition GPUs, while the local Windows host is reserved for manuscript compilation and artifact inspection. The preserved software environment uses Python 3.11.15, PyTorch 2.11.0+\texttt{cu128}, Transformers 4.57.6, Accelerate 1.14.0, Datasets 5.0.0, and project-native adapter code for all reported runs.

\textbf{Training protocol and hyperparameter settings.}
\method freezes the backbone and optimizes adapters in the q/k/v/o/gate/up/down projections, using rank $r=8$, $\alpha=16$, dropout 0.05, an Adam-style optimizer with learning rate $1\times10^{-5}$, a cosine schedule with 3\% warmup, and adapter-only restoration at rollback points.

\textbf{Benchmarks and evaluation.}
We evaluate on the same ten benchmark families as \msft---CommonsenseQA, OpenBookQA, AQUA-RAT, GSM8K, SciQ, ARC-Easy, HellaSwag, WinoGrande, BoolQ, and MedMCQA---under 5-shot greedy average accuracy \citep{talmor-etal-2019-commonsenseqa,mihaylov-etal-2018-suit,ling-etal-2017-program,cobbe2021trainingverifierssolvemath,welbl-etal-2017-crowdsourcing,clark2018thinksolvedquestionanswering,zellers-etal-2019-hellaswag,sakaguchi2020winogrande,clark-etal-2019-boolq,pmlr-v174-pal22a}. The headline metric is the unweighted average over these ten benchmark families.

\textbf{Reporting convention.}
Table~\ref{tab:main-results} is the main experimental table. It reports the headline ten-benchmark accuracy and peak epoch for each of the five evaluated backbones. We use it to compare \method with the corresponding \msft row under the same retained-backbone set. Table~\ref{tab:ablation} provides the matched Qwen2.5-3B ablation used to separate task scheduling, adapter-state rollback, and Weight ANL. These two in-text tables support the main claims without relying on supplementary material.

\begin{table*}[t]
\centering
\small
\renewcommand{\arraystretch}{1.15}
\setlength{\tabcolsep}{2.8pt}
\renewcommand{\tabularxcolumn}[1]{m{#1}}
\begin{tabularx}{0.96\textwidth}{@{}>{\centering\arraybackslash}m{0.17\textwidth}
>{\centering\arraybackslash}X
>{\centering\arraybackslash}X
>{\centering\arraybackslash}m{0.065\textwidth}
>{\centering\arraybackslash}m{0.052\textwidth}
>{\centering\arraybackslash}m{0.045\textwidth}@{}}
\toprule
\textbf{Compared Method} & \textbf{Task scheduling} & \textbf{Adapter state} & \textbf{Weight ANL} & \textbf{Acc. (\%)} & \textbf{Ep.} \\
\midrule
SFT & Fixed multi-task mixture & Full or task-agnostic training state & No & 72.10 & 5.00 \\
\mbox{SRO-\method} & \makecell{Precomputed hard exclusions\\from a single roll-out} & Adapter-only continuation & Yes & 72.50 & 4.25 \\
\mbox{Soft SRO-\method} & \makecell{Precomputed soft down-weighting\\after estimated task peaks} & Adapter-only continuation & Yes & 71.80 & 4.50 \\
\makecell{\method\ w/o\ Weight\\ANL\ (LoRA \msft)} & Iterative peak detection, exclusion, rollback, and continuation & Linear LoRA adapter state & No & 72.40 & 2.50 \\
\mbox{\method\ w/o rollback} & Iterative peak detection and exclusion, without reverting to the peak checkpoint & Weight-transformed adapter state & Yes & 72.70 & 1.50 \\
\method & Iterative peak detection, exclusion, rollback, and continuation & Weight-transformed adapter state & Yes & \textbf{73.30} & 0.50 \\
\bottomrule
\end{tabularx}
\caption{\textbf{Ablation study on Qwen2.5-3B.} Each row changes one scheduling or adapter-state component.}
\label{tab:ablation}
\end{table*}

\subsection{Main Results across Evaluated Backbones}

Across the five evaluated backbones, Table~\ref{tab:main-results} shows that \method reaches 61.36\% average accuracy, compared with 59.85\% for the \msft comparison row, a gain of 1.51 percentage points. AuroSFT leads the corresponding \msft entry on all five backbones. On Qwen2.5-7B, it reaches 77.50\% at epoch 2.50, compared with 77.45\% at epoch 2.50 for our \msft reproduction, a small advantage of 0.05 points. We therefore emphasize the consistent aggregate ordering rather than a large margin on every backbone.

Table~\ref{tab:main-results} reports the retained-backbone averages and task-group results. AuroSFT group-level accuracy and epoch values follow the same overall-best checkpoint reporting convention as the average-accuracy block. Qwen2.5-3B entries in the average-accuracy block use our reproduced measurements for SFT, Continual SFT, DynamixSFT, IES, and \msft. The Qwen2.5-7B \msft aggregate is also our reproduction. The Qwen2.5-7B \msft task-group cells remain reported source values because reproduced per-group logs are unavailable, and the non-AuroSFT Qwen2.5-3B task-group cells are reconstructed decompositions constrained to match the reproduced overall averages under the original 4/4/2 task-group weighting.

These results support the paper's main design claim. Overfitting-aware task scheduling can be paired with mergeable adapter-only training while improving the retained-backbone average. Because the system also includes scheduler, injection, and checkpoint choices, the matched ablations below separate the component effects.

\subsection{Ablation Study}
\label{sec:abl}

\paragraph{Setup.}
Following \msft, we compare against scheduler variants that approximate task-wise peak timing with less adaptive state. SRO-AuroSFT uses one initial roll-out to precompute the exclusion order, and Soft SRO-AuroSFT replaces hard exclusion with soft down-weighting after the estimated task budget is exhausted. We also remove AuroSFT-specific components. The w/o Weight ANL variant replaces the nonlinear weight-transformed adapter with a linear LoRA state, and w/o rollback keeps task exclusion but continues from the post-roll-out state. Together, these rows isolate the schedule, the rollback state, and the nonlinear adapter parameterization under one matched backbone.

\paragraph{Interpretation protocol.}
The ablation results provide component-removal evidence for the state-transfer thesis. Each row specifies both the task-state rule and the parameter state being carried forward. On Qwen2.5-3B, the full design performs best. SRO-AuroSFT and Soft SRO-AuroSFT trail \method, favoring iterative peak detection with rollback over a static or softened schedule. Removing Weight ANL lowers accuracy from 73.30\% to 72.40\%, and removing rollback also degrades accuracy, indicating that both the weight-transformed adapter and the checkpointed adapter state contribute beyond homogeneous SFT.

\subsection{Further Analysis}
\label{sec:further}

\begin{figure}[t]
    \centering
    \includegraphics[width=\linewidth]{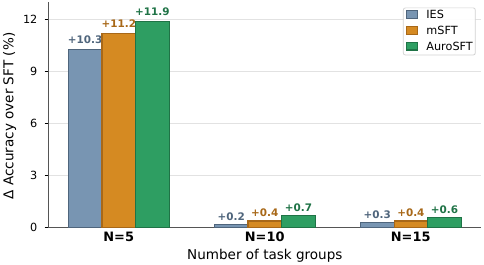}
    \caption{\textbf{Robustness across mixture scales on Qwen2.5-3B.} \method retains the largest gain over SFT at all evaluated mixture sizes under the matched setup.}
    \label{fig:aurosft_mixture_scale}
\end{figure}

\begin{figure*}[!t]
    \centering
    \includegraphics[width=0.98\textwidth]{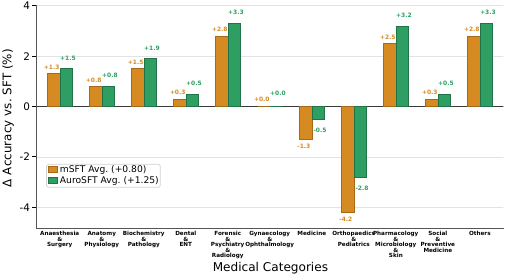}
    \vspace{-0.6em}
    \caption{\textbf{Fine-grained MedMCQA results on Qwen2.5-3B.} \method improves the reproduced average gain and reduces the largest negative category-level drops.}
    \label{fig:medmcqa-main}
\end{figure*}

\begin{figure}[!t]
    \centering
    \includegraphics[width=\linewidth]{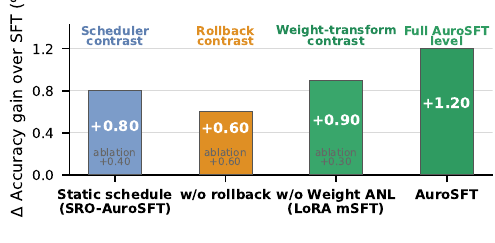}
    \caption{\textbf{Component sensitivity on Qwen2.5-3B.} Bars show component-removal contrasts relative to the full method under the matched ablation setup.}
    \label{fig:component-main}
\end{figure}

\begin{figure}[t]
    \centering
    \includegraphics[width=0.96\columnwidth]{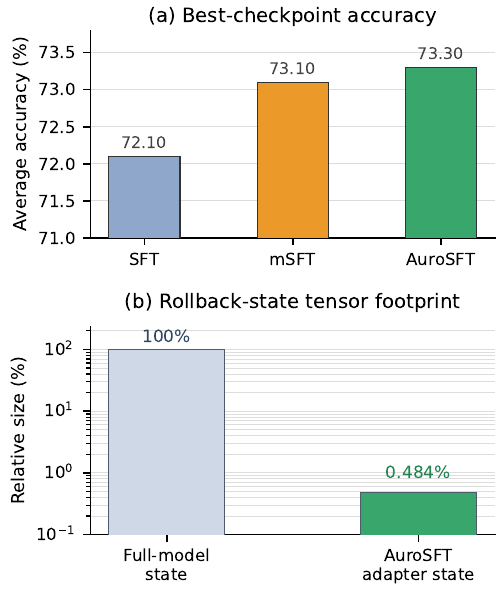}
    \vspace{-0.7em}
    \caption{\textbf{Accuracy and checkpoint-state efficiency on Qwen2.5-3B.} \method keeps the best checkpoint accuracy while sharply reducing the rollback-state tensor footprint.}
    \label{fig:efficiency-main}
\end{figure}

\begin{figure}[t]
    \centering
    \includegraphics[width=\columnwidth]{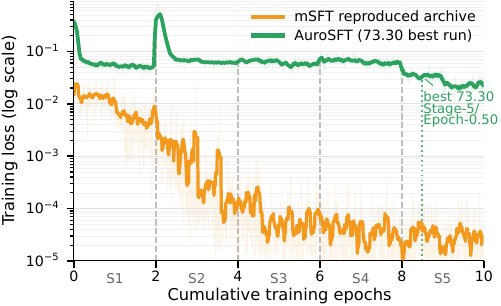}
    \caption{\textbf{Training-loss dynamics on Qwen2.5-3B.} The curves diagnose adapter-state continuation across stage boundaries in the matched run.}
    \label{fig:loss-dynamics-main}
\end{figure}

The main results and ablations evaluate final accuracy. We further examine when the framework helps, what state it changes, and how its training dynamics differ from full-state multi-task SFT. The compact diagnostics are shown in Figures~\ref{fig:aurosft_mixture_scale}, \ref{fig:medmcqa-main}, \ref{fig:component-main}, \ref{fig:efficiency-main}, and~\ref{fig:loss-dynamics-main}.

\paragraph{(I) Robustness across mixture scales.}
Following the scale-stress analysis in \msft, we vary the number of task groups on Qwen2.5-3B and measure improvements over a fixed-mixture SFT baseline. As shown in Figure~\ref{fig:aurosft_mixture_scale}, for $N\in\{5,10,15\}$ groups, AuroSFT obtains gains of $+11.9$, $+0.7$, and $+0.6$ points, respectively, compared with $+11.2$, $+0.4$, and $+0.4$ for reproduced \msft and $+10.3$, $+0.2$, and $+0.3$ for IES. The absolute margin is largest in the smaller mixture, but the ordering is stable across all three settings. This suggests that moving the scheduler state into adapters preserves the scale robustness of overfitting-aware scheduling while keeping the carried state compact.

\paragraph{(II) Fine-grained task granularity.}
We also revisit the MedMCQA granular decomposition used by \msft, where 21 medical subcategories are grouped into 11 broad categories. As shown in Figure~\ref{fig:medmcqa-main}, AuroSFT improves over SFT by $+1.25$ points on average across these medical groups, compared with $+0.80$ points for reproduced \msft. It is no worse than reproduced \msft on 10 of 11 groups and strictly better on 8. The largest negative categories are also less severe, suggesting that adapter-state continuation can reduce some negative transfer after tasks are excluded.

\paragraph{(III) Component sensitivity.}
The ablations in Table~\ref{tab:ablation} and Figure~\ref{fig:component-main} separate three effects that otherwise appear entangled. They isolate the schedule, the rollback operation, and the nonlinear weight transform. Single-roll-out and soft single-roll-out variants underperform the full method, so a static estimate of task peaks is not a sufficient substitute for iterative rollback. Removing rollback while keeping the adapter form also degrades accuracy, showing that the checkpointed state is an active part of the learning algorithm. Replacing Weight ANL with a linear LoRA state lowers accuracy from 73.30\% to 72.40\%, supporting the value of weight transformation beyond parameter efficiency alone.

\paragraph{(IV) Overfitting prevention and adapter-state forgetting.}
To make the task-level effect precise, let $\phi=\{A,B,\mathrm{ANL}\}$ denote the adapter state and define
\begin{equation}
\begin{aligned}
M_i(\phi)
&=\mathrm{Metric}\!\left(W_0+\Delta W(\phi);\mathcal D_i^{\mathrm{eval}}\right),\\
\Delta W(\phi)
&=\frac{\alpha}{r}B\bar A .
\end{aligned}
\end{equation}
If $\phi_i^\star$ is the adapter checkpoint at task $i$'s validation peak and $\phi_{\mathrm{final}}$ is the final adapter state, then the final gain over SFT can be decomposed as
\begin{equation}
\begin{aligned}
M_i(\phi_{\mathrm{final}})-M_i^{\mathrm{SFT}}
&=\big[M_i(\phi_i^\star)-M_i^{\mathrm{SFT}}\big]\\
&\quad+\big[M_i(\phi_{\mathrm{final}})-M_i(\phi_i^\star)\big].
\end{aligned}
\end{equation}
The first bracket captures overfitting prevention. The second captures forgetting or transfer after exclusion. The second term is negative when subsequent adapter continuation forgets an excluded task and positive when the remaining tasks transfer back to it. AuroSFT does not change this accounting identity. It changes the state space in which the identity is realized, replacing a full-model checkpoint with a compact, mergeable adapter state.

\paragraph{(V) Optimization dynamics.}
Figure~\ref{fig:loss-dynamics-main} plots Qwen2.5-3B training-loss traces for this state transition. The AuroSFT curve restarts from rolled-back adapter checkpoints at stage boundaries, while reproduced \msft tracks full-state training. We do not treat lower loss as proof of higher accuracy, since task conflicts affect which reductions matter. The curves instead show compact adapter-state continuation after rollback as exclusions reshape the active mixture.

\subsection{Efficiency and Deployment}

\method changes the resource profile of \msft. The original scheduler may store and restore full checkpoints at task peaks, whereas our implementation saves compact adapter artifacts and rollback metadata. Figure~\ref{fig:efficiency-main} shows that the AuroSFT adapter tensor state is 0.484\% of the corresponding full-model tensor state on Qwen2.5-3B. After training, $\Delta W=(\alpha/r)B\bar A$ merges into the frozen layer, preserving PEFT-style deployment while retaining \msft's task-state behavior.

\section{Discussion and Limitations}

\paragraph{Novelty and broader value.}
\method does not claim the discovery of heterogeneous overfitting, which comes from \msft, nor does it directly adopt input-level \aurora. Its new object is the \emph{stage-transfer state}. The scheduler carries a mergeable nonlinear low-rank weight parameterization rather than a full checkpoint, linking dynamic scheduling to the parameter object being transferred.

\paragraph{When the design should help.}
AuroSFT is most appropriate when tasks peak at different times, intermediate states must be retained, and the frozen backbone supports low-rank corrections. If task peaks are nearly aligned, a global stopping point may suffice. If stage states need not be recovered, adapter-only rollback offers less benefit. The design targets mixtures where heterogeneous task timing makes full-checkpoint transfer unnecessarily heavy.

\paragraph{Evidence scope and limitations.}
The evidence is strongest on five lightweight backbones and Qwen2.5-3B diagnostics, so we report system-level findings rather than a universal scaling law. Because the update remains input-linear and rank-bounded, we avoid claims from input-level \aurora theory. Stronger causal evidence requires matched multi-seed studies and profiling. Our efficiency claim is limited to rollback storage and mergeable deployment.

\paragraph{Auditable state trails.}
Each detected task peak maps to a compact adapter checkpoint, so exclusions, rollback choices, and final merged states can be audited without storing full backbones. The trail records which task triggered a transition, which adapter was restored, and which active mixture produced the next state, reducing attribution risk.

\paragraph{State-space design implication.}
This metadata clarifies which claims belong to the scheduler, the adapter state, or the overall system. Future adapter variants, rehearsal schemes, and rank-allocation policies can be compared by replacing the carried state under the same task-transition protocol.

\section{Conclusion}

We introduced \method, a parameter-efficient framework that combines overfitting-aware scheduling with nonlinear weight-transformed low-rank adapters. Its compact, mergeable stage-transfer state enables adapter-only rollback while remaining input-linear after merging. Across five lightweight backbones, \method improves the comparison-row average from 59.85\% to 61.36\% and leads on all five, while Qwen2.5-3B diagnostics support iterative scheduling, rollback, and Weight ANL. More broadly, the state carried by a dynamic data schedule is a design object. Task-specific timing can be preserved without full-model checkpoints.

\bibliography{aaai2027}

@article{grattafiori2024llama,
  title={The llama 3 herd of models},
  author={Grattafiori, Aaron and Dubey, Abhimanyu and Jauhri, Abhinav and Pandey, Abhinav and Kadian, Abhishek and Al-Dahle, Ahmad and Letman, Aiesha and Mathur, Akhil and Schelten, Alan and Vaughan, Alex and others},
  journal={arXiv preprint arXiv:2407.21783},
  year={2024}
}

@inproceedings{
walsh2025,
title={2 {OLM}o 2 Furious ({COLM}{\textquoteright}s Version)},
author={Evan Pete Walsh and Luca Soldaini and Dirk Groeneveld and Kyle Lo and Shane Arora and Akshita Bhagia and Yuling Gu and Shengyi Huang and Matt Jordan and Nathan Lambert and Dustin Schwenk and Oyvind Tafjord and Taira Anderson and David Atkinson and Faeze Brahman and Christopher Clark and Pradeep Dasigi and Nouha Dziri and Allyson Ettinger and Michal Guerquin and David Heineman and Hamish Ivison and Pang Wei Koh and Jiacheng Liu and Saumya Malik and William Merrill and Lester James Validad Miranda and Jacob Morrison and Tyler Murray and Crystal Nam and Jake Poznanski and Valentina Pyatkin and Aman Rangapur and Michael Schmitz and Sam Skjonsberg and David Wadden and Christopher Wilhelm and Michael Wilson and Luke Zettlemoyer and Ali Farhadi and Noah A. Smith and Hannaneh Hajishirzi},
booktitle={Second Conference on Language Modeling},
year={2025},
url={https://openreview.net/forum?id=2ezugTT9kU}
}

@misc{qwen2025qwen25technicalreport,
      title={Qwen2.5 Technical Report}, 
      author={Qwen and : and An Yang and Baosong Yang and Beichen Zhang and Binyuan Hui and Bo Zheng and Bowen Yu and Chengyuan Li and Dayiheng Liu and Fei Huang and Haoran Wei and Huan Lin and Jian Yang and Jianhong Tu and Jianwei Zhang and Jianxin Yang and Jiaxi Yang and Jingren Zhou and Junyang Lin and Kai Dang and Keming Lu and Keqin Bao and Kexin Yang and Le Yu and Mei Li and Mingfeng Xue and Pei Zhang and Qin Zhu and Rui Men and Runji Lin and Tianhao Li and Tianyi Tang and Tingyu Xia and Xingzhang Ren and Xuancheng Ren and Yang Fan and Yang Su and Yichang Zhang and Yu Wan and Yuqiong Liu and Zeyu Cui and Zhenru Zhang and Zihan Qiu},
      year={2025},
      eprint={2412.15115},
      archivePrefix={arXiv},
      primaryClass={cs.CL},
      url={https://arxiv.org/abs/2412.15115}, 
}

@article{shin2025dynamixsft,
  title={DynamixSFT: Dynamic Mixture Optimization of Instruction Tuning Collections},
  author={Shin, Haebin and Ji, Lei and Liu, Xiao and Yu, Zhiwei and Chen, Qi and Gong, Yeyun},
  journal={arXiv preprint arXiv:2508.12116},
  year={2025}
}

@article{JMLR:v23:21-0983,
  author  = {Ting Hu and Yunwen Lei},
  title   = {Early Stopping for Iterative Regularization with General Loss Functions},
  journal = {Journal of Machine Learning Research},
  year    = {2022},
  volume  = {23},
  number  = {339},
  pages   = {1--36},
  url     = {http://jmlr.org/papers/v23/21-0983.html}
}

@article{PRECHELT1998761,
title = {Automatic early stopping using cross validation: quantifying the criteria},
journal = {Neural Networks},
volume = {11},
number = {4},
pages = {761-767},
year = {1998},
issn = {0893-6080},
doi = {https://doi.org/10.1016/S0893-6080(98)00010-0},
url = {https://www.sciencedirect.com/science/article/pii/S0893608098000100},
author = {Lutz Prechelt}
}

@inproceedings{
yuan2025instancedependent,
title={Instance-dependent Early Stopping},
author={Suqin Yuan and Runqi Lin and Lei Feng and Bo Han and Tongliang Liu},
booktitle={The Thirteenth International Conference on Learning Representations},
year={2025},
url={https://openreview.net/forum?id=P42DbV2nuV}
}

@inproceedings{zhu-etal-2025-dynamic,
    title = "Dynamic Data Mixing Maximizes Instruction Tuning for Mixture-of-Experts",
    author = "Zhu, Tong  and
      Dong, Daize  and
      Qu, Xiaoye  and
      Ruan, Jiacheng  and
      Chen, Wenliang  and
      Cheng, Yu",
    editor = "Chiruzzo, Luis  and
      Ritter, Alan  and
      Wang, Lu",
    booktitle = "Proceedings of the 2025 Conference of the Nations of the Americas Chapter of the Association for Computational Linguistics: Human Language Technologies (Volume 1: Long Papers)",
    month = apr,
    year = "2025",
    address = "Albuquerque, New Mexico",
    publisher = "Association for Computational Linguistics",
    url = "https://aclanthology.org/2025.naacl-long.80/",
    doi = "10.18653/v1/2025.naacl-long.80",
    pages = "1663--1677",
    ISBN = "979-8-89176-189-6"
}

@inproceedings{dong-etal-2024-abilities,
    title = "How Abilities in Large Language Models are Affected by Supervised Fine-tuning Data Composition",
    author = "Dong, Guanting  and
      Yuan, Hongyi  and
      Lu, Keming  and
      Li, Chengpeng  and
      Xue, Mingfeng  and
      Liu, Dayiheng  and
      Wang, Wei  and
      Yuan, Zheng  and
      Zhou, Chang  and
      Zhou, Jingren",
    editor = "Ku, Lun-Wei  and
      Martins, Andre  and
      Srikumar, Vivek",
    booktitle = "Proceedings of the 62nd Annual Meeting of the Association for Computational Linguistics (Volume 1: Long Papers)",
    month = aug,
    year = "2024",
    address = "Bangkok, Thailand",
    publisher = "Association for Computational Linguistics",
    url = "https://aclanthology.org/2024.acl-long.12/",
    doi = "10.18653/v1/2024.acl-long.12",
    pages = "177--198"
}

@misc{shi2025damodatamixingoptimizer,
      title={DaMo: Data Mixing Optimizer in Fine-tuning Multimodal LLMs for Mobile Phone Agents}, 
      author={Kai Shi and Jun Yang and Ni Yang and Binqiang Pan and Qingsong Xie and Chao Zhang and Zhenyu Yang and Tianhuang Su and Haonan Lu},
      year={2025},
      eprint={2510.19336},
      archivePrefix={arXiv},
      primaryClass={cs.CV},
      url={https://arxiv.org/abs/2510.19336}, 
}

@inproceedings{
wang2026hbo,
title={{HBO}: Hierarchical Balancing Optimization for Fine-Tuning Large Language Models},
author={Weixuan Wang and Minghao Wu and Barry Haddow and Alexandra Birch},
booktitle={The Fourteenth International Conference on Learning Representations},
year={2026},
url={https://openreview.net/forum?id=JnhahbMvRE}
}

@inproceedings{li-etal-2024-quantity,
    title = "From Quantity to Quality: Boosting {LLM} Performance with Self-Guided Data Selection for Instruction Tuning",
    author = "Li, Ming  and
      Zhang, Yong  and
      Li, Zhitao  and
      Chen, Jiuhai  and
      Chen, Lichang  and
      Cheng, Ning  and
      Wang, Jianzong  and
      Zhou, Tianyi  and
      Xiao, Jing",
    editor = "Duh, Kevin  and
      Gomez, Helena  and
      Bethard, Steven",
    booktitle = "Proceedings of the 2024 Conference of the North American Chapter of the Association for Computational Linguistics: Human Language Technologies (Volume 1: Long Papers)",
    month = jun,
    year = "2024",
    address = "Mexico City, Mexico",
    publisher = "Association for Computational Linguistics",
    url = "https://aclanthology.org/2024.naacl-long.421/",
    doi = "10.18653/v1/2024.naacl-long.421",
    pages = "7602--7635"
}

@inproceedings{talmor-etal-2019-commonsenseqa,
    title = "{C}ommonsense{QA}: A Question Answering Challenge Targeting Commonsense Knowledge",
    author = "Talmor, Alon  and
      Herzig, Jonathan  and
      Lourie, Nicholas  and
      Berant, Jonathan",
    editor = "Burstein, Jill  and
      Doran, Christy  and
      Solorio, Thamar",
    booktitle = "Proceedings of the 2019 Conference of the North {A}merican Chapter of the Association for Computational Linguistics: Human Language Technologies, Volume 1 (Long and Short Papers)",
    month = jun,
    year = "2019",
    address = "Minneapolis, Minnesota",
    publisher = "Association for Computational Linguistics",
    url = "https://aclanthology.org/N19-1421/",
    doi = "10.18653/v1/N19-1421",
    pages = "4149--4158"
}

@inproceedings{mihaylov-etal-2018-suit,
    title = "Can a Suit of Armor Conduct Electricity? A New Dataset for Open Book Question Answering",
    author = "Mihaylov, Todor  and
      Clark, Peter  and
      Khot, Tushar  and
      Sabharwal, Ashish",
    editor = "Riloff, Ellen  and
      Chiang, David  and
      Hockenmaier, Julia  and
      Tsujii, Jun{'}ichi",
    booktitle = "Proceedings of the 2018 Conference on Empirical Methods in Natural Language Processing",
    month = oct # "-" # nov,
    year = "2018",
    address = "Brussels, Belgium",
    publisher = "Association for Computational Linguistics",
    url = "https://aclanthology.org/D18-1260/",
    doi = "10.18653/v1/D18-1260",
    pages = "2381--2391"
}

@inproceedings{ling-etal-2017-program,
    title = "Program Induction by Rationale Generation: Learning to Solve and Explain Algebraic Word Problems",
    author = "Ling, Wang  and
      Yogatama, Dani  and
      Dyer, Chris  and
      Blunsom, Phil",
    editor = "Barzilay, Regina  and
      Kan, Min-Yen",
    booktitle = "Proceedings of the 55th Annual Meeting of the Association for Computational Linguistics (Volume 1: Long Papers)",
    month = jul,
    year = "2017",
    address = "Vancouver, Canada",
    publisher = "Association for Computational Linguistics",
    url = "https://aclanthology.org/P17-1015/",
    doi = "10.18653/v1/P17-1015",
    pages = "158--167"
}

@misc{cobbe2021trainingverifierssolvemath,
      title={Training Verifiers to Solve Math Word Problems}, 
      author={Karl Cobbe and Vineet Kosaraju and Mohammad Bavarian and Mark Chen and Heewoo Jun and Lukasz Kaiser and Matthias Plappert and Jerry Tworek and Jacob Hilton and Reiichiro Nakano and Christopher Hesse and John Schulman},
      year={2021},
      eprint={2110.14168},
      archivePrefix={arXiv},
      primaryClass={cs.LG},
      url={https://arxiv.org/abs/2110.14168}, 
}

@inproceedings{welbl-etal-2017-crowdsourcing,
    title = "Crowdsourcing Multiple Choice Science Questions",
    author = "Welbl, Johannes  and
      Liu, Nelson F.  and
      Gardner, Matt",
    editor = "Derczynski, Leon  and
      Xu, Wei  and
      Ritter, Alan  and
      Baldwin, Tim",
    booktitle = "Proceedings of the 3rd Workshop on Noisy User-generated Text",
    month = sep,
    year = "2017",
    address = "Copenhagen, Denmark",
    publisher = "Association for Computational Linguistics",
    url = "https://aclanthology.org/W17-4413/",
    doi = "10.18653/v1/W17-4413",
    pages = "94--106"
}

@misc{clark2018thinksolvedquestionanswering,
      title={Think you have Solved Question Answering? Try ARC, the AI2 Reasoning Challenge}, 
      author={Peter Clark and Isaac Cowhey and Oren Etzioni and Tushar Khot and Ashish Sabharwal and Carissa Schoenick and Oyvind Tafjord},
      year={2018},
      eprint={1803.05457},
      archivePrefix={arXiv},
      primaryClass={cs.AI},
      url={https://arxiv.org/abs/1803.05457}, 
}

@inproceedings{zellers-etal-2019-hellaswag,
    title = "{H}ella{S}wag: Can a Machine Really Finish Your Sentence?",
    author = "Zellers, Rowan  and
      Holtzman, Ari  and
      Bisk, Yonatan  and
      Farhadi, Ali  and
      Choi, Yejin",
    editor = "Korhonen, Anna  and
      Traum, David  and
      M{\`a}rquez, Llu{\'i}s",
    booktitle = "Proceedings of the 57th Annual Meeting of the Association for Computational Linguistics",
    month = jul,
    year = "2019",
    address = "Florence, Italy",
    publisher = "Association for Computational Linguistics",
    url = "https://aclanthology.org/P19-1472/",
    doi = "10.18653/v1/P19-1472",
    pages = "4791--4800"
}

@inproceedings{sakaguchi2020winogrande,
  title={WinoGrande: An Adversarial Winograd Schema Challenge at Scale},
  author={Sakaguchi, Keisuke and Le Bras, Ronan and Bhagavatula, Chandra and Choi, Yejin},
  booktitle={Proceedings of the AAAI Conference on Artificial Intelligence},
  volume={34},
  pages={8732--8740},
  year={2020}
}

@inproceedings{clark-etal-2019-boolq,
    title = "{B}ool{Q}: Exploring the Surprising Difficulty of Natural Yes/No Questions",
    author = "Clark, Christopher  and
      Lee, Kenton  and
      Chang, Ming-Wei  and
      Kwiatkowski, Tom  and
      Collins, Michael  and
      Toutanova, Kristina",
    editor = "Burstein, Jill  and
      Doran, Christy  and
      Solorio, Thamar",
    booktitle = "Proceedings of the 2019 Conference of the North {A}merican Chapter of the Association for Computational Linguistics: Human Language Technologies, Volume 1 (Long and Short Papers)",
    month = jun,
    year = "2019",
    address = "Minneapolis, Minnesota",
    publisher = "Association for Computational Linguistics",
    url = "https://aclanthology.org/N19-1300/",
    doi = "10.18653/v1/N19-1300",
    pages = "2924--2936"
}

@InProceedings{pmlr-v174-pal22a,
  title = 	 {MedMCQA: A Large-scale Multi-Subject Multi-Choice Dataset for Medical domain Question Answering},
  author =       {Pal, Ankit and Umapathi, Logesh Kumar and Sankarasubbu, Malaikannan},
  booktitle = 	 {Proceedings of the Conference on Health, Inference, and Learning},
  pages = 	 {248--260},
  year = 	 {2022},
  editor = 	 {Flores, Gerardo and Chen, George H and Pollard, Tom and Ho, Joyce C and Naumann, Tristan},
  volume = 	 {174},
  series = 	 {Proceedings of Machine Learning Research},
  month = 	 {07--08 Apr},
  publisher =    {PMLR},
  url = 	 {https://proceedings.mlr.press/v174/pal22a.html}
}

@article{li2025data,
  title={Data mixing optimization for supervised fine-tuning of large language models},
  author={Li, Yuan and Liu, Zhengzhong and Xing, Eric},
  journal={arXiv preprint arXiv:2508.11953},
  year={2025}
}

@article{xiao2024sftmix,
  title={Sftmix: Elevating language model instruction tuning with mixup recipe},
  author={Xiao, Yuxin and Zhang, Shujian and Zhou, Wenxuan and Ghassemi, Marzyeh and Zhao, Sanqiang},
  journal={arXiv preprint arXiv:2410.05248},
  year={2024}
}

@article{koh2026msft,
  title={mSFT: Addressing Dataset Mixtures Overfitting Heterogeneously in Multi-task SFT},
  author={Koh, Woosung and Jeon, Jeyoung and Song, Youngjin and Cheon, Yujin and Oh, Soowon and Choi, Jaehyeong and Yun, Se-Young},
  journal={arXiv preprint arXiv:2603.21606},
  year={2026}
}

@article{dong2025aurora,
  title={AuroRA: Breaking Low-Rank Bottleneck of LoRA with Nonlinear Mapping},
  author={Dong, Haonan and Zhu, Wenhao and Song, Guojie and Wang, Liang},
  journal={arXiv preprint arXiv:2505.18738},
  year={2025}
}

@article{hu2021lora,
  title={LoRA: Low-Rank Adaptation of Large Language Models},
  author={Hu, Edward J. and Shen, Yelong and Wallis, Phillip and Allen-Zhu, Zeyuan and Li, Yuanzhi and Wang, Shean and Wang, Lu and Chen, Weizhu},
  journal={arXiv preprint arXiv:2106.09685},
  year={2021}
}

@inproceedings{houlsby2019parameter,
  title={Parameter-Efficient Transfer Learning for NLP},
  author={Houlsby, Neil and Giurgiu, Andrei and Jastrzebski, Stanislaw and Morrone, Bruna and de Laroussilhe, Quentin and Gesmundo, Andrea and Attariyan, Mona and Gelly, Sylvain},
  booktitle={Proceedings of the 36th International Conference on Machine Learning},
  year={2019}
}

@article{li2021prefix,
  title={Prefix-Tuning: Optimizing Continuous Prompts for Generation},
  author={Li, Xiang Lisa and Liang, Percy},
  journal={arXiv preprint arXiv:2101.00190},
  year={2021}
}

@inproceedings{lester2021power,
  title={The Power of Scale for Parameter-Efficient Prompt Tuning},
  author={Lester, Brian and Al-Rfou, Rami and Constant, Noah},
  booktitle={Proceedings of the 2021 Conference on Empirical Methods in Natural Language Processing},
  year={2021}
}

@article{dettmers2023qlora,
  title={QLoRA: Efficient Finetuning of Quantized LLMs},
  author={Dettmers, Tim and Pagnoni, Artidoro and Holtzman, Ari and Zettlemoyer, Luke},
  journal={arXiv preprint arXiv:2305.14314},
  year={2023}
}

@article{zhang2023adalora,
  title={AdaLoRA: Adaptive Budget Allocation for Parameter-Efficient Fine-Tuning},
  author={Zhang, Qingru and Chen, Minshuo and Bukharin, Alexander and Karampatziakis, Nikos and He, Pengcheng and Cheng, Yu and Chen, Weizhu and Zhao, Tuo},
  journal={arXiv preprint arXiv:2303.10512},
  year={2023}
}

@article{zhang2023instructiontuning,
  title={Instruction Tuning for Large Language Models: A Survey},
  author={Zhang, Shengyu and Dong, Linfeng and Li, Xiaoya and Zhang, Sen and Sun, Xiaofei and Wang, Shuhe and Li, Jiwei and Hu, Runyi and Zhang, Tianwei and Wu, Fei and Wang, Guoyin},
  journal={arXiv preprint arXiv:2308.10792},
  year={2023}
}

@article{liu2024dora,
  title={DoRA: Weight-Decomposed Low-Rank Adaptation},
  author={Liu, Shih-Yang and Wang, Chien-Yi and Yin, Hongxu and Molchanov, Pavlo and Wang, Yu-Chiang Frank and Cheng, Kwang-Ting and Chen, Min-Hung},
  journal={arXiv preprint arXiv:2402.09353},
  year={2024}
}

@article{wang2024loraga,
  title={LoRA-GA: Low-Rank Adaptation with Gradient Approximation},
  author={Wang, Shaowen and Yu, Linxi and Li, Jian},
  journal={arXiv preprint arXiv:2407.05000},
  year={2024}
}
\end{document}